%% file: main.tex
\documentclass[runningheads]{llncs}

\usepackage{eccv}

\usepackage{eccvabbrv}

\usepackage{graphicx}
\usepackage{booktabs}
\usepackage{multirow}
\usepackage[table]{xcolor}
\usepackage{pifont}
\usepackage{makecell}
\usepackage{array}
\usepackage{adjustbox}
\usepackage{soul}

\newcommand{\cmark}{\ding{51}}
\newcommand{\xmark}{\textcolor{red}{\ding{55}}}
\newcolumntype{L}[1]{...}
\newcolumntype{R}[1]{...}

\usepackage[accsupp]{axessibility}  

\input{preamble}

\usepackage{hyperref}

\usepackage{orcidlink}

\begin{document}

\title{WeatherReasonSeg: A Benchmark for Weather-Aware Reasoning Segmentation in Visual Language Models} 

\titlerunning{WeatherReasonSeg}

\author{Wanjun Du\inst{1}$^*$ \and
Zifeng Yuan\inst{2}\thanks{Equal Contribution} \and
Tingting Chen\inst{2} \and
Fucai Ke\inst{3} \and
\\
Beibei Lin\inst{2}\textsuperscript{$\dagger$} \and
Shunli Zhang\inst{1}\textsuperscript{$\ddagger$}
}

\def\thefootnote{$\ddagger$}\footnotetext{Corresponding Author}
\def\thefootnote{$\dagger$}\footnotetext{Project Leader}

\authorrunning{W.~Du et al.}

\institute{Beijing Jiaotong University \and
National University of Singapore \and
Monash University \\
\email{\{25110603, slzhang\}@bjtu.edu.cn, \{zyuan, tingting.c, beibei.lin\}@u.nus.edu, fucai.ke1@monash.edu}}

\maketitle

\begin{abstract}

Existing vision-language models (VLMs) have demonstrated impressive performance in reasoning-based segmentation. 
However, current benchmarks are primarily constructed from high-quality images captured under idealized conditions. This raises a critical question: when visual cues are severely degraded by adverse weather conditions such as rain, snow, or fog, can VLMs sustain reliable reasoning segmentation capabilities? 
In response to this challenge, we introduce WeatherReasonSeg, a benchmark designed to evaluate VLM performance in reasoning-based segmentation under adverse weather conditions. 
It consists of two complementary components. First, we construct a controllable reasoning dataset by applying synthetic weather with varying severity levels to existing segmentation datasets, enabling fine-grained robustness analysis. Second, to capture real-world complexity, we curate a real-world adverse-weather reasoning segmentation dataset with semantically consistent queries generated via mask-guided LLM prompting. We further broaden the evaluation scope across five reasoning dimensions, including functionality, application scenarios, structural attributes, interactions, and requirement matching.
Extensive experiments across diverse VLMs reveal two key findings: (1) VLM performance degrades monotonically with increasing weather severity, and (2) different weather types induce distinct vulnerability patterns. We hope WeatherReasonSeg will serve as a foundation for advancing robust, weather-aware reasoning.

\end{abstract}

\begin{figure*}[!h]
    \centering
    \includegraphics[width=1\linewidth]{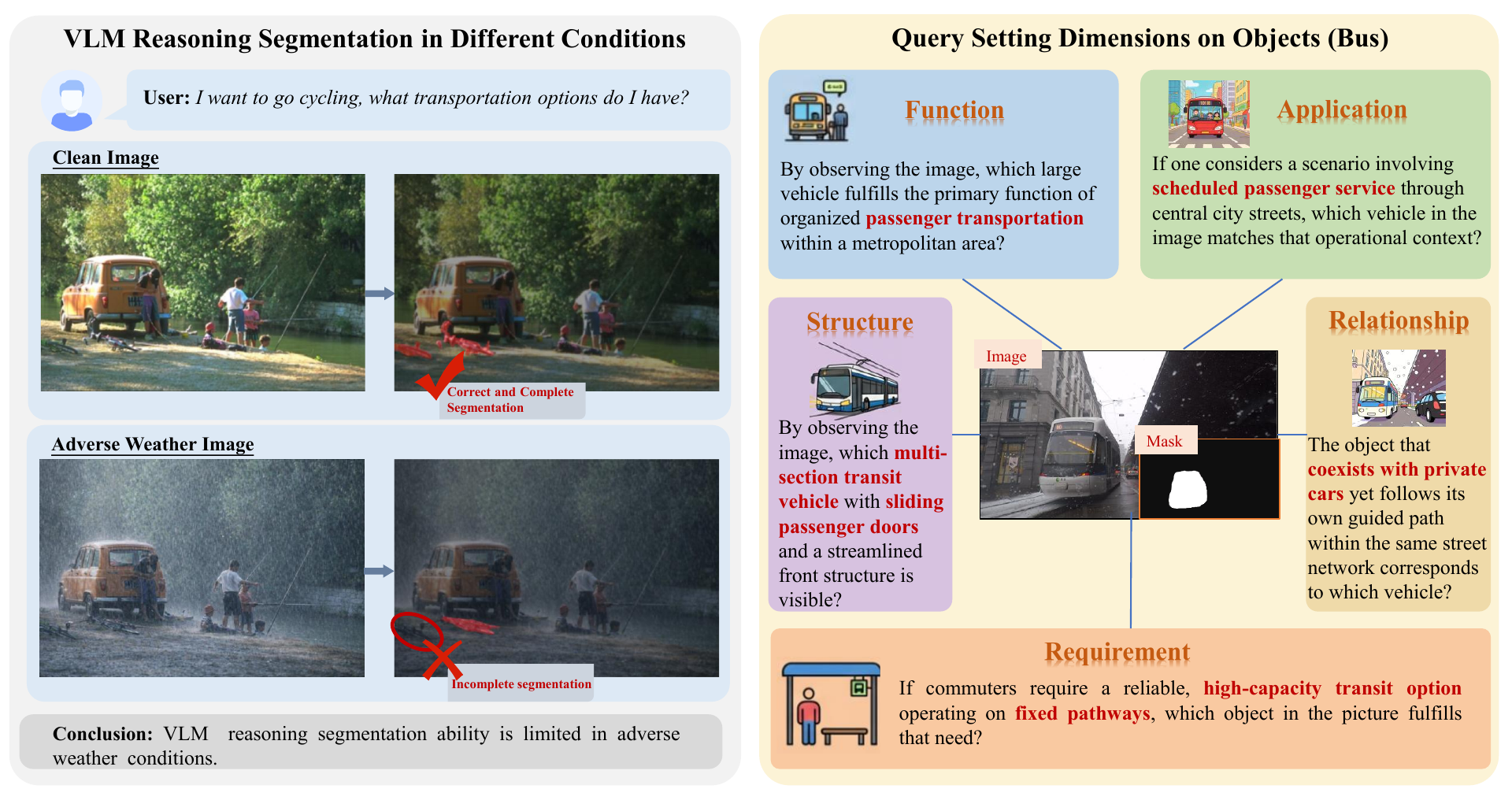}
    \caption{\textbf{Adverse weather undermines VLM reasoning segmentation.}
\textbf{Left:} Under clean weather conditions, VLMs can correctly localize the target object and generate complete segmentation masks. However, when exposed to adverse weather, the model fails to accurately ground the reasoning query, resulting in incomplete segmentation masks and degraded pixel-level alignment and semantic reasoning performance. 
\textbf{Right:} WeatherReasonSeg addresses this limitation by introducing real-world degraded data with five structured reasoning dimensions—Function, Application, Structure, Relationship, and Requirement—to enable systematic evaluation of weather-robust reasoning.}
    \label{fig_teaser} 
\end{figure*}




\section{Introduction}
\label{sec:intro}

Recent advances in Vision-Language Models (VLMs) have significantly improved multimodal reasoning tasks, including reasoning-based segmentation~\cite{liu2023visual,wahed2024prima,liu2025seg}. Representative benchmarks such as ReasonSeg~\cite{lai2024lisa} enable models to generate segmentation masks conditioned on complex language descriptions rather than simple object labels. However, existing evaluations are conducted under ideal visual conditions, implicitly assuming high-quality imagery.

In real-world outdoor environments, adverse weather conditions such as fog, rain, and snow frequently degrade visual quality by reducing visibility, distorting structures, and introducing noise~\cite{lin2024nightrain, chen2024dual}. These degradations disrupt visual-language alignment and can propagate through multi-stage reasoning pipelines, leading to amplified downstream errors in mask prediction \cite{zhang2025mamba, yang2025erf, yang2025semantic, yang2024end, yang2024semantic}. Despite its practical importance in safety-critical deployment, systematic evaluation of reasoning-based segmentation under adverse weather remains largely unexplored.

To bridge this gap, we introduce WeatherReasonSeg, a pioneering benchmark specifically designed to evaluate reasoning-based segmentation performance under adverse weather conditions. 
WeatherReasonSeg comprises two complementary components: a controllable adverse-weather dataset and a real-world adverse-weather dataset. 
In the former, we synthesize adverse weather effects of different types and severity levels on existing reasoning-based segmentation datasets, covering graded interference conditions including light, moderate, and severe fog, rain, and snow. 
This controlled setting enables systematic exploration of the reasoning limits of VLMs under varying interference intensities. Concurrently, we construct a real-world dataset to ensure the authenticity of adverse conditions, thereby enabling more accurate quantification of reasoning degradation under real visual corruption.

Furthermore, we formulate contextual queries across five critical dimensions: target function, application scenario, structural attributes, interaction relationships, and requirement matching. We adopt a mask-guided large language model prompting strategy to generate queries aligned with real images, enabling VLMs to perform context-aware reasoning and facilitating comprehensive evaluation across diverse reasoning dimensions. 

Experimental results reveal a sharp deterioration in VLM performance under severe weather conditions. In synthetic benchmarks, the average accuracy of VLMs decreased by 15\% compared to ideal environments; while in real-world scenarios, the accuracy of reasoning-based methods only reached half of the upper limit of perception-based methods, indicating that the reasoning stage becomes the main bottleneck. These findings quantitatively reveal the vulnerability of current VLMs to the environment and highlight the necessity of developing controllable, severity-aware benchmarks like WeatherReasonSeg to drive the development of weather-robust reasoning models.

Our contributions are summarized as follows:
\begin{itemize}
\item We propose the first benchmark, \textbf{WeatherReasonSeg}, specifically designed to evaluate the reasoning capabilities of VLMs under adverse weather conditions, bridging the gap between idealized benchmark evaluation and real-world deployment scenarios.

\item We construct a synthetic dataset with graded weather interference, covering multiple weather types and severity levels, which quantitatively characterizes the impact of weather intensity on reasoning processes and segmentation performance.

\item We build a real-world reasoning-based segmentation dataset and adopt a mask-guided large language model prompting strategy to generate semantically consistent, executable, and highly diverse queries. These queries encourage VLMs to perform complex compositional reasoning, extending beyond the task scope of traditional reasoning-based segmentation.
\end{itemize}









\section{Related Work}
\textbf{Reasoning in Large Models} Recent years have witnessed substantial advances in the reasoning capabilities of Large Language Models (LLMs)~\cite{ke2025explain, chen2025hypospace,chen2025auto}. OpenAI-o1~\cite{openaio1} shows that extending Chain-of-Thought (CoT) reasoning during inference, which is often referred to as inference-time scaling, can significantly improve complex reasoning. Building upon this paradigm, subsequent works explore test-time scaling strategies via process-based reward modeling~\cite{wang2023mathshepherd,uesato2022solvingmath,lightman2023stepbystep}, reinforcement learning (RL) optimization~\cite{kumar2024training,shao2024deepseekmath}, and search-based inference mechanisms~\cite{trinh2024solving,feng2023alphazerolike}. Among these approaches, DeepSeek-R1~\cite{guo2025deepseekr1} employs the GRPO algorithm~\cite{shao2024deepseekmath} and achieves strong reasoning performance with only a few thousand RL training steps, highlighting the efficiency of RL-based policy optimization for enhancing reasoning ability.

Inspired by these advances in LLM reasoning, recent efforts have begun transferring reasoning-oriented training paradigms to multimodal large language models (MLLMs). For example, Open-R1-Multimodal~\cite{openr1multimodal} focuses on multimodal mathematical reasoning, while R1-V~\cite{r1v} demonstrates improvements in visual counting tasks. However, existing multimodal reasoning methods (\eg, ~\cite{ke2025dwim, suris2023vipergpt, openr1multimodal, ke2024hydra, r1v}) primarily focus on high-level semantic or symbolic reasoning and are typically evaluated on benchmarks such as OK-VQA~\cite{marino2019okvqa}, A-OKVQA~\cite{schwenk2022aokvqa}, and VCR~\cite{zellers2019vcr_dataset}, which involve clear and well-curated images. As a result, these approaches seldom address fine-grained pixel-level understanding or reasoning under visually degraded conditions.
In contrast, our work concentrates on pixel-level reasoning for visual perception and investigates reinforcement learning as a principled mechanism to bridge reasoning and dense segmentation. 

\textbf{Semantic Segmentation with Reasoning} Semantic segmentation aims to perform dense pixel-wise classification by assigning a semantic category label to each pixel. Extensive prior research~\cite{long2015fully,chen2017deeplab,lin2017refinenet, chen2017rethinking, zhao2017pyramid,badrinarayanan2017segnet, cheng2021per,ronneberger2015u, zhang2024heap, zhang2023adaptive}, has driven significant progress in this area. Representative models such as DeepLab~\cite{chen2018encoder}, MaskFormer~\cite{cheng2022masked}, and Segment Anything Model (SAM)~\cite{kirillov2023sam,ravi2024sam} establish strong and stable baselines, making conventional category-driven segmentation a relatively mature problem.

To overcome the limitations of predefined label spaces, referring expression segmentation (RES) \cite{kazemzadeh2014referitgame,yu2016refcoco} introduces natural language descriptions to specify target regions. In this setting, models must align short textual expressions with corresponding visual entities. LISA \cite{lai2024lisa} further extends this framework to reasoning-based segmentation, where queries may be longer, more abstract, or require multi-step reasoning. Such tasks demand deeper joint reasoning over linguistic semantics and visual cues to accurately localize and segment the intended object.

\textbf{MLLMs for Segmentation} Following the introduction of the <SEG> token in LISA \cite{lai2024lisa,yang2023lisa++}, which enables interaction between multimodal large language models (MLLMs) and segmentation architectures, a number of subsequent works \cite{chen2024sam4mllm,ren2024pixellm,bai2024one} explore integrating MLLMs into segmentation pipelines. Many of these methods, including OneTokenSegAll \cite{bai2024one} and PixelLM \cite{ren2024pixellm}, adopt token-based designs that connect language models with segmentation decoders via special interface tokens.

While effective, such tightly coupled frameworks typically require large-scale annotated data to jointly fine-tune both the MLLM and the segmentation module. Moreover, additional token-level interactions may disturb the original pixel-level representations learned by pretrained segmentation networks. In contrast, our approach adopts a decoupled design that preserves the structural integrity of existing segmentation models while leveraging the reasoning capability of MLLMs to enhance segmentation performance.

\section{Methodology}

\begin{figure*}[t!]
    \centering
    \includegraphics[width=1\linewidth]{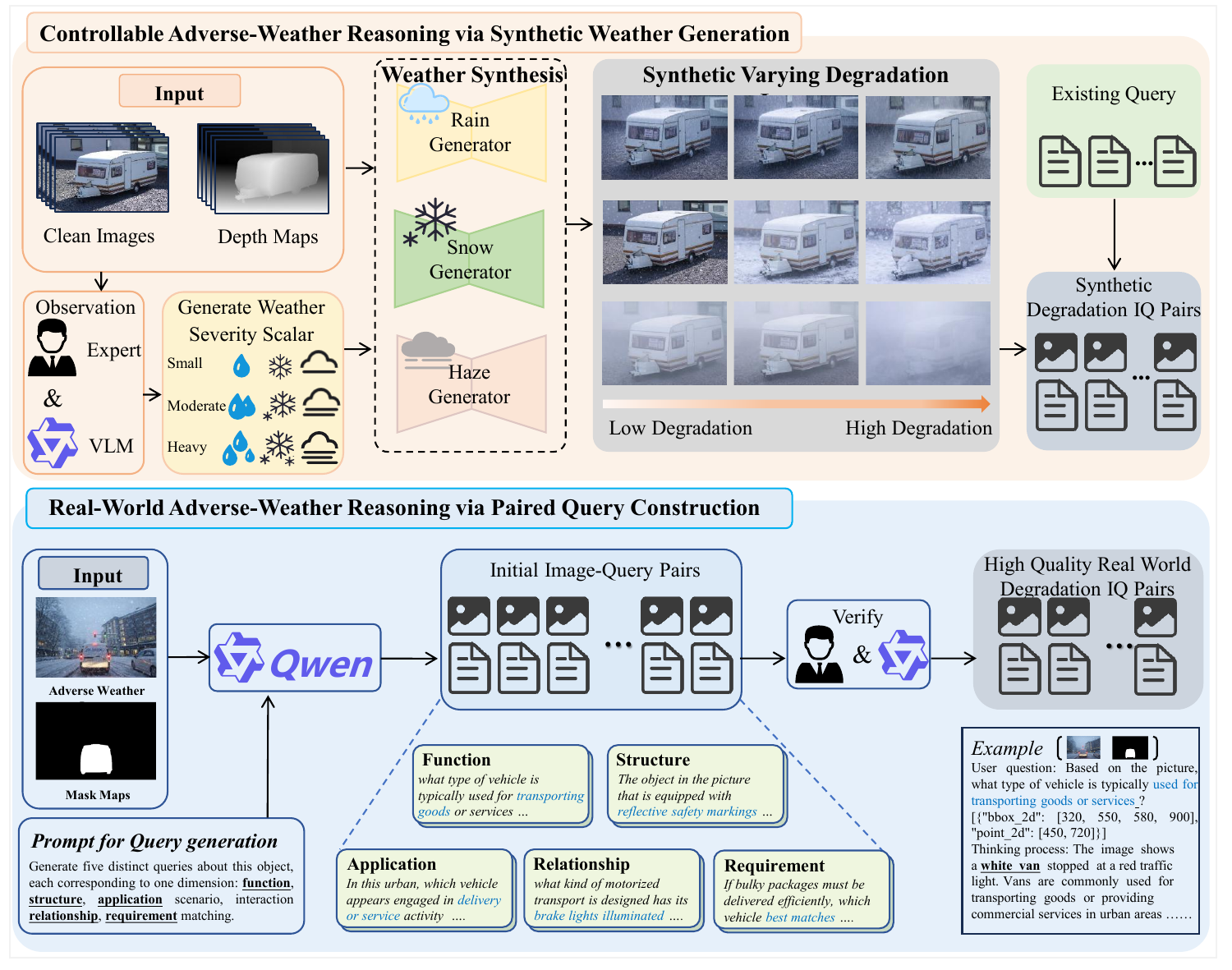}
    \caption{Overview of WeatherReasonSeg. The framework consists of two complementary components. 
    \textbf{Top:} A controllable synthetic weather generation process can generate weather type degradations (rain, snow, haze) of varying severity to construct synthetic image-query pairs for robustness evaluation. 
    \textbf{Bottom:} A real-world adverse-weather reasoning dataset constructed via mask-guided large language model prompting, followed by human–model collaborative verification to ensure semantic alignment and reasoning validity. }
    \label{fig_pipeline}
\end{figure*}

Figure~\ref{fig_pipeline} illustrates the overall pipeline of WeatherReasonSeg, which comprises two complementary components: a controllable adverse-weather reasoning dataset and a real-world adverse-weather reasoning dataset. 
The first dataset synthesizes different weather degradations across three severity levels based on existing reasoning datasets~\cite{lai2024lisa}, providing a controlled environment to evaluate model stability under varying conditions. 
Complementarily, the second dataset leverages a mask-guided LLM prompting to construct high-quality, real-world image-query pairs, capturing the diverse reasoning dimensions inherent in actual adverse weather.

\subsection{Controllable Adverse-Weather Reasoning via Synthetic Weather Generation}

To enable a systematic robustness evaluation of VLM reasoning capabilities, we adopt an existing reasoning-based segmentation dataset as the foundation and synthesize controllable weather degradations on top of it. This synthetic strategy allows explicit manipulation of weather types and severity levels (e.g., fog density, rainfall intensity, and snow accumulation), thereby supporting fine-grained analysis of how progressively increasing visual disturbances affect reasoning stability.

\textbf{Base Dataset} 
ReasonSeg~\cite{lai2024lisa} is the first and most representative benchmark specifically designed for reasoning-driven segmentation. Unlike conventional semantic segmentation or referring expression datasets, ReasonSeg emphasizes implicit reasoning without relying on explicit category labels, requiring models to infer targets through attributes, relationships, and contextual constraints.

\textbf{Adverse Weather Synthesis Process} Building upon the need for controllable degradation modeling, we introduce a physically interpretable weather synthesis mechanism into the ReasonSeg dataset, forming the synthetic component of our benchmark-WeatherReasonSeg. The design follows two core principles. First, the synthesis process should follow real imaging physics rather than simple noise injection, which ensures realistic degradation effects. Second, the degradation severity must be continuously controllable. This allows systematic analysis of model performance under different weather types and levels. 
The overall pipeline is illustrated in Figure~\ref{fig_pipeline}. Clean images and their corresponding depth maps are fed into dedicated weather generators for rain, snow, and fog. Based on a weather severity scalar derived from expert knowledge and VLM-assisted calibration, different levels of degradation are applied to the images. The degraded images are then paired with the original reasoning queries to construct degraded image–query pairs.

Specifically, we model three representative weather types, namely rain, snow, and fog, according to their underlying physical characteristics, incorporating effects such as rain streak accumulation, snowflake scattering and occlusion, and atmospheric light scattering. The severity of each weather type is continuously controlled by adjusting physically meaningful parameters, such as rain density, snow particle distribution, and visibility distance, thereby generating degradations ranging from light to severe conditions. 
Overall, this physics-guided and controllable synthesis framework enables structured and fine-grained modeling of adverse weather across multiple types and severity levels, facilitating systematic analysis of how environmental disturbances influence VLM reasoning stability. Detailed implementation of the synthesis process and specific parameter settings for each weather generator are provided in the Supplementary Material.

\begin{table*}[t!]
\centering
\caption{Comparison of recent benchmarks and resources for vision-language reasoning and segmentation. We evaluate each dataset by whether it provides bounding boxes and segmentation masks, supports general reasoning and adverse-weather settings, and by its data type and amount.}
\label{tab:dataset_comparison}
\scriptsize
\setlength{\tabcolsep}{3pt}
\renewcommand{\arraystretch}{1.15}
\begin{adjustbox}{max width=.98\linewidth}
\begin{tabular}{
    >{\raggedright\arraybackslash}p{4.2cm}
    c c c c c
    >{\raggedleft\arraybackslash}p{1.1cm}
}
\toprule
Dataset (Venue, Year)
& \makecell{Bounding\\Boxes}
& \makecell{Seg.\\Masks}
& \makecell{General\\Reasoning}
& \makecell{Adverse\\Weather}
& \makecell{Data\\Type}
& Amount \\
\midrule

RainCityscapes (CVPR'18)\cite{hu2019depth}     & \xmark & \cmark & \xmark & \cmark & Syn         & 10.6k \\
ACDC (ICCV'21)\cite{sakaridis2025acdc}         & \cmark & \cmark & \xmark & \cmark & Real        & 4k \\
Rain Wcity (IJCAI'22)\cite{ijcai2022p243}      & \xmark & \cmark & \xmark & \cmark & Real        & 24.3k \\
CREPE (CVPR'23)\cite{ma2023crepe}              & \cmark & \xmark & \cmark & \xmark & --          & 1162.6k \\
LVIS-Ground (ECCV'24)\cite{ma2024groma}        & \xmark & \xmark & \cmark & \xmark & --          & 4.3k \\
M4-Instruct (CVPR'24)\cite{li2024llava}        & \xmark & \xmark & \cmark & \xmark & --          & 307k \\
FP-RefCOCO (CVPR'24)\cite{wu2024see}           & \cmark & \cmark & \xmark & \xmark & --          & 65.8k \\
LLM-Seg40K (CVPR'24)\cite{wang2024llm}         & \cmark & \cmark & \xmark & \xmark & --          & 40k \\
MRES-32M (CVPR'24)\cite{wang2024unveiling}     & \cmark & \cmark & \xmark & \xmark & --          & 32M \\
NaturalBench (CVPR'24)\cite{li2024naturalbench}& \xmark & \xmark & \cmark & \xmark & --          & 10k \\
VAB (CVPR'25)\cite{hsu2025makes}               & \xmark & \xmark & \cmark & \xmark & --          & 0.54k \\
ReasonSeg (CVPR'24)\cite{lai2024lisa}          & \cmark & \cmark & \cmark & \xmark & --          & 1.2k \\
\midrule

\rowcolor[gray]{0.88}
\textbf{WeatherReasonSeg (ours)} & \cmark & \cmark & \cmark & \cmark & Syn \& Real & \textbf{44.7k} \\
\bottomrule
\end{tabular}
\end{adjustbox}
\end{table*}

\subsection{Real-World Adverse-Weather Reasoning via Paired Query Construction}

Although physics-based synthetic data can systematically control the type and severity of image degradation, it cannot fully capture the complexity of real-world conditions. 
Therefore, we further construct a reasoning-based segmentation dataset for real-world adverse weather conditions. This dataset uses real-world degraded images while preserving high-quality pixel-level annotations. Based on this, we employ a mask-guided large-scale language model prompting mechanism to generate multidimensional natural language reasoning queries.

\textbf{Base Dataset}
Firstly, we adopt ACDC (Adverse Conditions Dataset with Correspondences)~\cite{SDV21} as the foundation for real-world degraded scenarios. It covers multiple adverse weather conditions, including rain, fog, snow, and low-light (nighttime) environments. A key advantage of ACDC is its high-quality pixel-level annotations and cross-weather consistency, making it well-suited for studying the impact of visual degradation on perception and reasoning. Unlike synthetic data, the degradations arise from real physical environments and imaging processes, including complex occlusions, non-uniform scattering, and changes in semantic visibility.

Based on the original segmentation annotations, we extract target masks and construct reasoning-driven language queries aligned with the pixel-level regions, extending ACDC into a real-world degraded dataset tailored for visual-language reasoning segmentation tasks.

\textbf{Query Generation}
To systematically evaluate the reasoning capabilities of VLMs in realistic degradation scenarios, we employ a mask-guided prompting strategy to automatically generate natural language queries that are strictly aligned with the target segmentation region. Furthermore, we pose questions about the same target object from multiple reasoning perspectives to characterize the differences in VLM's reasoning behavior across different contextual dimensions. As illustrated in Figure \ref{fig_pipeline}, adverse weather images and the corresponding extracted mask maps are input into a large language model. Guided by carefully designed prompts, the model generates queries across five different reasoning dimensions, forming initial image-query pairs. These pairs are then jointly validated and filtered by human annotators and the large language model according to predefined rules. The final output consists of high-quality real-world degradation image-query pairs.

Unlike queries that only describe object categories or simple attributes~\cite{lai2024lisa}, we design queries from five key reasoning dimensions: 
(1) \emph{Function}, focusing on the intrinsic purpose of the object; (2) \emph{Application Scenario}, emphasizing contextual usage environments; (3) \emph{Structural/Feature}, targeting observable attributes and compositional details; (4) \emph{Relational}, examining spatial or interactive relationships with surrounding entities; and (5) \emph{Requirement Matching}, mapping practical needs to appropriate objects.
Illustrative examples of these five reasoning dimensions are provided in Figures~\ref{fig_teaser} and~\ref{fig_pipeline}. 

\textbf{Query Dataset Filtering}
Beyond careful image selection and prompt design, we implement a rigorous filtering strategy to ensure query quality and semantic consistency. This process eliminates query–answer pairs with potential ambiguity or task misalignment. Specifically: (a) We discard questions that merely enumerate objects or components without functional or contextual semantics, avoiding shallow descriptive queries. (b) We remove question groups that fail to maintain single-target consistency, where all five questions do not explicitly refer to the same physical object. (c) We retain only question–answer groups in which all questions refer to the same explicit and visually identifiable named entity, ensuring semantic clarity and disambiguation. (d) We exclude question–answer pairs that mention objects absent from the image, visually unclear, or semantically ambiguous, ensuring all queries can be reliably answered based solely on the input image.

\subsection{Dataset Statistics and Analysis}

\begin{figure*}[!h]
    \centering
    \includegraphics[width=0.98\linewidth]{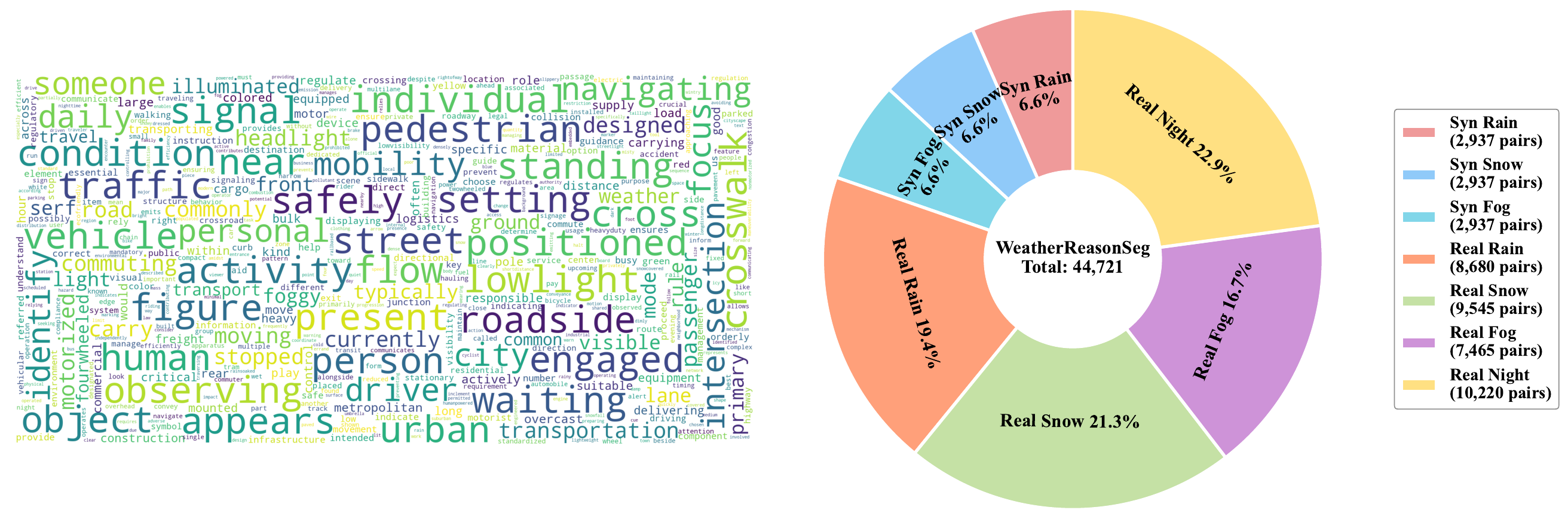}
    \caption{Overview of WeatherReasonSeg. 
Left: Query word cloud. 
Right: Distribution of query–image pairs across weather conditions.}
    \label{fig_data} 
\end{figure*}

Our WeatherReasonSeg contains a total of 44,721 image-query pairs spanning both synthetic and real-world environments. The synthetic subset includes 2,937 image–query pairs for each weather type (rain, snow, and fog), amounting to 8,811 pairs for controlled robustness evaluation. The real-world subset further enhances ecological validity, comprising 8,680 rainy, 9,545 snowy, 7,465 foggy, and 10,220 nighttime pairs, totaling 35,910 pairs. As illustrated in Fig.~\ref{fig_data}, the dataset provides a diverse distribution of query semantics and weather conditions. Compared with prior reasoning segmentation benchmarks (e.g., ReasonSeg with 1.2k samples), it substantially enlarges the evaluation scale while introducing structured environmental perturbations, providing a comprehensive benchmark for assessing VLM robustness under adverse conditions.

Table \ref{tab:dataset_comparison} provides multi-dimensional comparisons between our WeatherReasonSeg and other relevant benchmarks.
As shown in the table, WeatherReasonSeg stands as the only benchmark that simultaneously supports bounding boxes, pixel-level masks, general reasoning supervision, and explicit adverse-weather settings. Existing weather-oriented datasets focus on perception under degraded conditions but lack reasoning queries, while recent reasoning-centric benchmarks assume clean visual environments and do not model environmental disturbances. WeatherReasonSeg bridges this gap by unifying pixel-grounded reasoning and diverse adverse-weather scenarios within a single framework.



\section{Experiments}
\subsection{Implementation Details}
\paragraph{\textbf{Baselines}}
We evaluate four representative frameworks on WeatherReasonSeg: Grounded-SAM~\cite{ren2024grounded}, which is grounding-based, and three reasoning-based models: LISA~\cite{lai2024lisa}, Seg-R1~\cite{you2025seg}, and Seg-Zero~\cite{liu2025seg}.
%
%

\begin{table*}[t]
\centering
\caption{Performance on WeatherReasonSeg under different weather types and severities. Seg-Zero* is the reasoning-only model without SAM (bbox evaluation).}
\label{tab:weatherreasonseg_giou_ciou}
\setlength{\tabcolsep}{1.8pt}
\renewcommand{\arraystretch}{1}
\footnotesize
\begin{adjustbox}{max width=.98\linewidth}
\begin{tabular}{lcccccccccccc}
\toprule
\multirow{3}{*}{Method}
& \multicolumn{4}{c}{Fog}
& \multicolumn{4}{c}{Rain}
& \multicolumn{4}{c}{Snow} \\
\cmidrule(lr){2-5}\cmidrule(lr){6-9}\cmidrule(lr){10-13}
& \multicolumn{2}{c}{val} & \multicolumn{2}{c}{test}
& \multicolumn{2}{c}{val} & \multicolumn{2}{c}{test}
& \multicolumn{2}{c}{val} & \multicolumn{2}{c}{test} \\
\cmidrule(lr){2-3}\cmidrule(lr){4-5}
\cmidrule(lr){6-7}\cmidrule(lr){8-9}
\cmidrule(lr){10-11}\cmidrule(lr){12-13}
& gIoU & cIoU & gIoU & cIoU
& gIoU & cIoU & gIoU & cIoU
& gIoU & cIoU & gIoU & cIoU \\
\midrule

\rowcolor[gray]{0.88}
\multicolumn{13}{c}{\textbf{Clean}} \\
Grounded SAM        & 26.0    & 14.5 & 21.3 & 16.4 & 26.0    & 14.5 & 21.3 & 16.4 & 26.0    & 14.5 & 21.3 & 16.4 \\
LISA-7B             & 53.6  & 52.3 & 48.7 & 48.8 & 53.6  & 52.3 & 48.7 & 48.8 & 53.6  & 52.3 & 48.7 & 48.8 \\
Seg-R1              & 60.8  & 56.2 & 55.3 & 46.6 & 60.8  & 56.2 & 55.3 & 46.6 & 60.8  & 56.2 & 55.3 & 46.6 \\
Seg-Zero*  & 65.3 & --   & 59.3& --   & 65.3 & --   & 59.3& --   & 65.3 & --   & 59.3& --   \\
Seg-Zero-7B         & 62.6  & 62.0 & 57.5 & 52.0 & 62.6  & 62.0 & 57.5 & 52.0 & 62.6  & 62.0 & 57.5 & 52.0 \\

\midrule
\rowcolor[gray]{0.88}
\multicolumn{13}{c}{\textbf{Severity: Light}} \\
Grounded SAM        & 23.4 & 12.3 & 18.5 & 14.3 & 22.2 & 13.6 & 16.4 & 15.3 & 22.9 & 13.0 & 16.2 & 15.0 \\
LISA-7B             & 44.3 & 50.4 & 41.8 & 47.2 & 44.5 & 49.1 & 41.4 & 47.1 & 41.5 & 48.2 & 41.0 & 47.5 \\
Seg-R1              & 48.4 & 44.3 & 46.9 & 40.8 & 48.5 & 41.5 & 43.9 & 42.5 & 40.1 & 36.9 & 37.7 & 35.8 \\
Seg-Zero*  & 58.4 & --   & 55.9 & --   & 59.7 & --   & 56.1 & --   & 53.7 & --   & 55.0 & --   \\
Seg-Zero-7B         & 57.1 & 57.4 & 53.9 & 48.4 & 58.6 & 56.8 & 54.0 & 48.6 & 53.2 & 56.8 & 53.1 & 47.0 \\

\midrule
\rowcolor[gray]{0.88}
\multicolumn{13}{c}{\textbf{Severity: Moderate}} \\
Grounded SAM        & 19.9 & 14.2 & 16.1 & 13.9 & 17.3 & 12.9 & 14.8 & 13.4 & 16.1 & 12.3 & 13.5 & 12.9 \\
LISA-7B             & 41.8 & 48.1 & 40.7 & 46.2 & 42.3 & 47.2 & 39.2 & 46.6 & 38.9 & 43.5 & 37.4 & 46.1 \\
Seg-R1              & 50.8 & 46.6 & 47.8 & 41.9 & 52.9 & 48.1 & 46.6 & 43.5 & 40.8 & 38.6 & 44.2 & 44.3 \\
Seg-Zero*  & 58.2 & --   & 53.2 & --   & 57.9 & --   & 53.7 & --   & 52.3 & --   & 47.0 & --   \\
Seg-Zero-7B         & 56.0 & 57.3 & 50.9 & 45.3 & 56.6 & 56.1 & 51.5 & 47.0 & 52.0 & 50.3 & 45.7 & 44.5 \\

\midrule
\rowcolor[gray]{0.88}
\multicolumn{13}{c}{\textbf{Severity: Heavy}} \\
Grounded SAM        & 15.8 & 11.2 & 12.8 & 12.3 & 14.3 & 10.8 & 12.2 & 11.8 & 13.9 & 11.6 & 11.7 & 10.8 \\
LISA-7B             & 40.7 & 47.7 & 39.4 & 48.4 & 40.1 & 46.9 & 38.7 & 45.2 & 38.2 & 42.9 & 36.6 & 44.3 \\
Seg-R1              & 46.4 & 35.5 & 43.4 & 36.5 & 42.3 & 39.3 & 41.5 & 40.3 & 37.1 & 31.1 & 35.4 & 34.9 \\
Seg-Zero*  & 57.5 & --   & 51.7 & --   & 54.4 & --   & 50.5 & --   & 47.7 & --   & 48.0 & --   \\
Seg-Zero-7B         & 55.4 & 53.4 & 49.9 & 44.0 & 54.3 & 54.3 & 47.8 & 47.7 & 45.9 & 49.7 & 45.7 & 39.7 \\

\bottomrule
\end{tabular}
\end{adjustbox}
\end{table*}

\paragraph{\textbf{Evaluation Metrics}} Following previous works\cite{kazemzadeh2014referitgame,yu2016refcoco}, we calculate gIoU and cIoU. The gIoU is the average of all per-image Intersection-over-Unions (IoUs), while the cIoU calculates the cumulative intersection over the cumulative union. Unless specified, we use gIoU as our default metric, as it equally considers both large and small objects.

\subsection{Results on Synthesis Segmentation Benchmarks}
We first evaluate model performance on the WeatherReasonSeg-Synthesis benchmark, where weather degradations (rain, fog, and snow) are progressively introduced at three severity levels (severity: Light/Moderate/Heavy). This controlled setting enables quantitative assessment of how reasoning robustness degrades with increasing weather intensity.

\begin{figure*}[t!]
    \centering
    \includegraphics[width=1\linewidth]{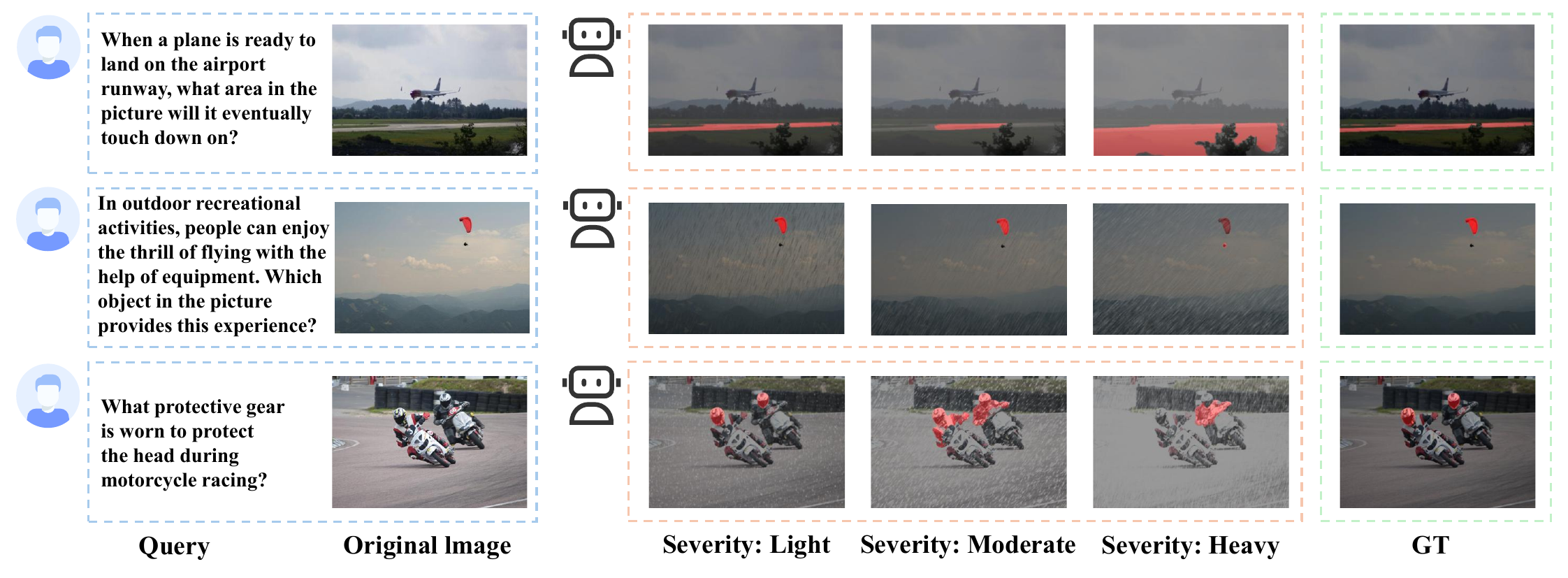}
    \caption{A schematic comparison of VLM reasoning segmentation results under different weather degradation levels. The VLM-based segmentation method used is Seg-Zero. The figure illustrates how segmentation performance progressively changes as weather severity increases.. }
    \label{fig_exper}
\end{figure*}

As shown in Figure~\ref{tab:weatherreasonseg_giou_ciou}, all models perform well under clean conditions without weather interference. However, as weather severity increases from mild (severity: Light) to severe (severity: Heavy), all methods show a clear and consistent decline in both gIoU and cIoU. This indicates that adverse weather introduces systematic disruptions to VLMs reasoning. 
Notably, in the Seg-Zero-7B* setting, which involves no additional segmentation optimization and thus directly reflects the VLM’s reasoning quality, the gIoU decreases from 65.3 in clean conditions to 47.7 under severe weather, representing an absolute drop of 17.6.
This directly shows that weather-induced visual degradation weakens the VLM’s semantic association and spatial reasoning ability.

The full Seg-Zero-7B model also exhibits substantial performance drops. In foggy conditions, the gIoU decreases from 57.5 to 49.9, and in snowy scenes, the cIoU declines from 52 to 39.72. Grounded-SAM achieved a cIoU of only 10.8 in heavy fog. The impact of weather types is not uniform. Snow leads to the most severe degradation, followed by fog and then rain. This pattern is consistent with the strong loss of texture, contrast, and depth cues in snowy scenes.

Figure~\ref{fig_exper} further provides qualitative examples of Seg-Zero under increasing weather severity. Under light weather interference, the model can still produce relatively accurate segmentation results. However, as weather conditions progressively deteriorate, the segmentation quality declines noticeably. In moderately degraded scenarios, foggy and snowy scenes already exhibit incomplete object localization and spatial drift. Under severe degradation, all weather types show significant segmentation errors or incomplete predictions. These observations suggest that the severe loss of visual cues caused by adverse weather significantly limits the segmentation reasoning capability of vision-language models.

\subsection{Results on Real World Segmentation Benchmarks}
To further evaluate VLM reasoning performance under real-world adverse weather conditions, we conduct experiments on the WeatherReasonSeg-RealWorld benchmark, which contains naturally degraded images captured under fog, rain, snow, and nighttime conditions. We first measure the upper bound of perception robustness using the pure segmentation model SAM2, and then evaluate VLM-based reasoning and segmentation methods under the same settings. As shown in ~\cref{tab:acdc_realworld}, SAM2 achieves strong and stable performance across all conditions, with gIoU scores of 82.2 (fog), 80.1 (rain), 81.5 (snow), and 74.8 (night). Although adverse weather slightly reduces accuracy, the overall performance remains high, establishing a robust perception upper bound.

In contrast, VLM-driven reasoning segmentation methods exhibit substantial performance degradation. The best-performing Seg-Zero-7B achieves only 40.6 gIoU and 47.1 cIoU in fog, and further drops to 28.9 gIoU and 29.4 cIoU at night—approximately half of SAM2’s performance under identical conditions. Grounded-SAM performs even worse, with cIoU falling below 15\% at night. These results indicate a clear performance gap between perception-only and reasoning-based models in real-world degraded environments.

\begin{table*}[t]
\centering
\caption{Performance comparison on real-world adverse weather. 
Gray rows indicate model grouping.SAM2 serves as the performance upper bound because it is directly prompted with ground-truth spatial locations (e.g., bounding boxes).The overall best performances are shown in \textbf{bold}, while the second best performances are shown \underline{underlined}.}
\label{tab:acdc_realworld}
\setlength{\tabcolsep}{6pt}
\renewcommand{\arraystretch}{1}
\small
\begin{adjustbox}{max width=.98\linewidth}
\begin{tabular}{lcccccccc}
\toprule
\textbf{Method} 
& \multicolumn{2}{c}{fog}
& \multicolumn{2}{c}{Rain}
& \multicolumn{2}{c}{Snow}
& \multicolumn{2}{c}{Night} \\
\cmidrule(lr){2-3}\cmidrule(lr){4-5}\cmidrule(lr){6-7}\cmidrule(lr){8-9}
& gIoU$\uparrow$ & cIoU$\uparrow$ & gIoU$\uparrow$ & cIoU$\uparrow$ & gIoU$\uparrow$ & cIoU$\uparrow$ & gIoU$\uparrow$ & cIoU$\uparrow$ \\
\midrule

\rowcolor[gray]{0.9}
\multicolumn{9}{c}{\textbf{SAM2 (Upper Bound)}} \\

SAM2 
& 82.2 & 91.6 
& 80.1 & 87.5 
& 81.5 & 89.5 
& 74.8 & 84.4 \\

\midrule

\rowcolor[gray]{0.9}
\multicolumn{9}{c}{\textbf{Reasoning and Segmentation}} \\

Grounded SAM 
& 23.7 & 14.4 
& 18.8 & 16.5 
& 15.3 & 14.8 
& 11.4 & 12.1 \\

LISA-7B 
& 18.9 & 24.4 
& 11.5 & 18.1 
& 14.9 & 23.6
& 14.7 & 25.3 \\

Seg-R1 
& \underline{30.9} & \underline{44.4} 
& \underline{26.7} & \underline{36.1} 
& \underline{29.4} & \underline{37.9} 
& \underline{18.5} & \underline{26.3} \\

Seg-Zero-7B 
& \textbf{40.6} & \textbf{47.1} 
& \textbf{35.1} & \textbf{38.7} 
& \textbf{36.8} & \textbf{41.6} 
& \textbf{28.9} & \textbf{29.4} \\

\bottomrule
\end{tabular}
\end{adjustbox}
\end{table*}

\begin{figure*}[t!]
    \centering
    \includegraphics[width=1\linewidth]{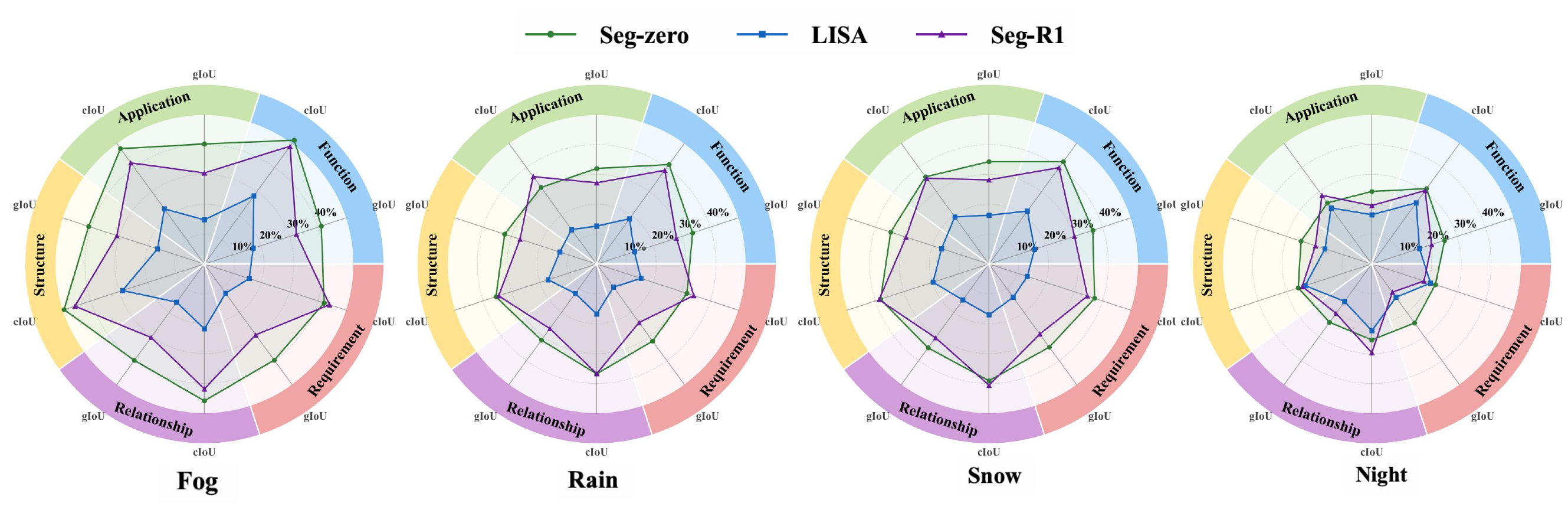}
    \caption{Radar chart comparison of reasoning segmentation performance across five reasoning dimensions under different weather conditions (Fog, Rain, Snow, and Night). Results are reported using gIoU and cIoU for three representative VLM-based methods: Seg-Zero, LISA, and Seg-R1. The figure shows that visually grounded dimensions such as Structure and Function remain relatively stable, while context-dependent dimensions including Application and Requirement exhibit larger performance degradation under adverse weather conditions. }
    \label{fig_radar}
\end{figure*}

Next, we analyze model performance across five semantic query dimensions: Function, Application, Structure, Relationship, and Requirement. As shown in Figure~\ref{fig_radar} and summarized in Figure~\ref{tab:dimension_analysis}, consistent trends appear across all weather conditions. The Structure and Function dimensions consistently achieve higher segmentation performance, while Application and Requirement show noticeably lower accuracy. For example, under rainy conditions, Seg-Zero achieves a cIoU of 41.2 for Function queries but only 31.8 for Requirement queries, indicating a nearly 10\% gap. This pattern suggests that queries grounded in directly observable visual attributes remain relatively robust, whereas those requiring higher-level contextual reasoning suffer greater degradation under adverse weather.

\begin{table*}[t]
\centering
\caption{Reasoning performance across five query dimensions under different weather conditions.
The five dimensions correspond to Function, Application, Structure, Relationship, and Requirement.The overall best performances are shown in \textbf{bold}, while the second best performances are shown \underline{underlined}.}
\label{tab:dimension_analysis}
\setlength{\tabcolsep}{8pt}
\renewcommand{\arraystretch}{1}
\small
\begin{adjustbox}{max width=.98\linewidth}
\begin{tabular}{llcccccc}
\toprule
Weather & Dimension 
& \multicolumn{2}{c}{Seg-Zero}
& \multicolumn{2}{c}{LISA}
& \multicolumn{2}{c}{Seg-R1} \\
\cmidrule(lr){3-4}\cmidrule(lr){5-6}\cmidrule(lr){7-8}
& & gIoU$\uparrow$ & cIoU$\uparrow$ & gIoU$\uparrow$ & cIoU$\uparrow$ & gIoU$\uparrow$ & cIoU$\uparrow$ \\
\midrule

\multirow{5}{*}{Fog}
& Function      & \textbf{41.2} & \textbf{51.2} & \textbf{17.2} & \underline{28.2} & \textbf{32.4} & \textbf{48.8} \\
& Application   & 40.2 & 47.8 & 14.8 & 22.8 & 30.5 & 41.9 \\
& Structure     & \underline{40.8} & \underline{49.4} & 16.5 & \textbf{28.8} & \underline{30.8} & \underline{45.6} \\
& Relationship  & 39.9 & 45.8 & 15.8 & 21.7 & 30.4 & 41.8 \\
& Requirement   & 39.8 & 42.2 & 12.1 & 15.8 & 29.3 & 44.1 \\
\midrule

\multirow{5}{*}{Rain}
& Function      & 33.8 & 41.2 & 13.3 & 18.7 & 28.1 & 38.8 \\
& Application   & 32.0 & 31.7 & 12.7 & 14.3 & 27.2 & 36.2 \\
& Structure     & 32.4 & 35.5 & 13.0 & 17.1 & 27.0 & 34.5 \\
& Relationship  & 31.5 & 36.8 & 12.2 & 16.7 & 26.7 & 36.7 \\
& Requirement   & 31.9 & 31.8 &  9.6 & 15.6 & 24.2 & 34.2 \\
\midrule

\multirow{5}{*}{Snow}
& Function      & 36.6 & 42.4 & 16.1 & 21.9 & 30.1 & 39.9 \\
& Application   & 34.3 & 36.1 & 16.3 & 19.5 & 28.2 & 35.5 \\
& Structure     & 34.6 & 38.2 & 16.7 & 19.7 & 29.2 & 38.7 \\
& Relationship  & 34.7 & 39.1 & 14.9 & 17.0 & 30.5 & 40.6 \\
& Requirement   & 34.4 & 37.2 & 13.8 & 13.5 & 29.0 & 34.7 \\
\midrule

\multirow{5}{*}{Night}
& Function      & 25.6 & 31.2 & \underline{16.8} & 25.3 & 21.1 & 30.4 \\
& Application   & 24.3 & 25.3 & 16.5 & 23.2 & 19.6 & 28.4 \\
& Structure     & 24.9 & 25.9 & 16.5 & 23.5 & 19.9 & 24.5 \\
& Relationship  & 24.1 & 25.4 & 15.5 & 22.3 & 20.4 & 29.7 \\
& Requirement   & 24.4 & 22.4 & 13.8 & 20.7 & 11.7 & 18.4 \\
\bottomrule
\end{tabular}
\end{adjustbox}
\end{table*}

\subsection{Discussions}
\textbf{Impact of Weather Severity on Reasoning Stability} Figure ~\ref{tab:weatherreasonseg_giou_ciou} shows that weather severity acts as a graded stress test for VLM-driven reasoning segmentation. While all models perform strongly under clean conditions, performance consistently declines as severity increases from mild to severe across both gIoU and cIoU. This monotonic degradation indicates that adverse weather systematically disrupts pixel-grounded reasoning.

The Seg-Zero-7B* setting highlights this instability. Since this configuration reflects pure reasoning quality, the gIoU drop from 65.3 to 47.7 under severe weather directly reveals that visual degradation weakens semantic association and spatial grounding.  Grounded-SAM nearly collapses under heavy fog. These results suggest that low-level visual corruption propagates to high-level reasoning, causing structural instability in the pipeline.

\textbf{Impact of Real-World Adverse Weather on VLM Reasoning}
Even under real-world adverse conditions, the performance gap between perception-only and reasoning-based models remains substantial. As shown in Figure~\ref{tab:acdc_realworld}, SAM2 maintains relatively strong segmentation accuracy across fog, rain, snow, and nighttime scenarios, whereas all reasoning-based methods experience significant performance degradation under the same conditions. Notably, across all weather types, reasoning and segmentation models fail to close the gap with SAM2, indicating that the primary bottleneck does not originate from low-level feature extraction. Instead, instability arises at the semantic reasoning stage, where degraded visual inputs disrupt reliable visual-semantic alignment before segmentation begins.

Moreover, the impact of different weather conditions is not uniform. Nighttime and fog lead to the most severe performance drops, while rain causes comparatively smaller degradation. This suggests that environments involving strong illumination loss or reduced global visibility more severely impair the visual cues required for reliable visual-semantic reasoning.

\textbf{Performance Across Different Semantic Reasoning Dimensions} Figure~\ref{tab:dimension_analysis} reveals substantial differences in VLMs' reasoning performance across the five semantic query dimensions, highlighting an internal imbalance in their reasoning capabilities. Structure- and Function-oriented queries consistently outperform Application- and Requirement-based queries. This discrepancy arises because Application and Requirement queries rely more heavily on contextual abstraction and scenario-level reasoning, rather than directly observable visual attributes. As a result, they are more sensitive to degraded environmental cues.

\textbf{Insights for Future Research} Our benchmark provides several important implications for future research. First, although VLM-driven reasoning segmentation achieves strong performance under ideal conditions, VLMs exhibit significant degradation under real-world adverse weather. The primary bottleneck lies in the absence of weather-aware reasoning mechanisms capable of modeling uncertainty and adapting semantic grounding under distribution shift. Second, the observed performance differences across weather types and semantic reasoning dimensions suggest that future benchmarks should incorporate a broader range of environmental scenarios and higher-level reasoning tasks to better evaluate robustness. Finally, since current VLMs lack explicit modeling of severe weather conditions, future research could explore integrating physically grounded weather simulation models into pretraining or fine-tuning pipelines. By incorporating degradation-aware supervision, VLMs may learn to extract more reliable semantic cues from visually corrupted inputs.

\section{Conclusion}
In this paper, we introduce WeatherReasonSeg, the first benchmark for systematically evaluating reasoning-based segmentation of vision-language models (VLMs) under adverse weather conditions. The benchmark includes two complementary components: a controllable synthetic weather dataset with multiple weather types and severity levels for robustness analysis, and a real-world adverse-weather reasoning dataset constructed via mask-guided LLM prompting to ensure semantically consistent and spatially grounded queries. To comprehensively assess reasoning behavior, we further organize the queries across five semantic dimensions: Function, Application, Structure, Relationship, and Requirement.Extensive experiments reveal the limitations of existing reasoning-based segmentation approaches and provide a comprehensive performance analysis of this underexplored problem. Our results show that current VLMs lack environment-adaptive reasoning mechanisms, leading to substantial performance degradation under adverse weather. We hope this benchmark will serve as a foundation for advancing weather-aware and robust reasoning systems in safety-critical real-world applications.



%
%
\bibliographystyle{splncs04}
\bibliography{main}

\clearpage
\appendix

\begin{center}
    {\LARGE \boldmath \bfseries Supplementary Material \par}
    \vspace{2em}
\end{center}

\section{Weather Synthesis with Details}

We introduce physically interpretable weather synthesis mechanisms in WeatherReasonSeg to systematically model degradation on the original images of ReasonSeg. Our design follows two key principles: (1) the synthesis process should approximate real-world imaging mechanisms under adverse weather, rather than relying on heuristic noise injection; and (2) the degradation severity should be continuously controllable to enable quantitative analysis across different weather types and intensity levels.

Based on these principles, we construct physically consistent synthetic data with different severity levels by modeling three representative weather conditions: rain, fog, and snow.

\subsection{Adverse Weather Synthesis}

\subsubsection{Rain Synthesis}

Rain degradation mainly manifests as two visual effects: sparse rain streaks and accumulation effects caused by dense overlapping streaks. To model these characteristics, we adopt a region dependent rain model and further incorporate rain accumulation~\cite{yang2017deep, lin2024nightrain, chen2024dual, teng2025raindropgs}. 
The basic rain model is formulated as:
\begin{equation}
O = B + S R,
\end{equation}
where $O$ denotes the observed rainy image, $B$ is the clean background image, $S$ represents the rain streak intensity layer, and $R \in \{0,1\}$ is a region dependent binary mask indicating rain affected pixels. This formulation decouples the spatial distribution of rain streaks from their intensity, enabling structured modeling of rain patterns.

The above model mainly describes sparse rain conditions. Under heavy rainfall, dense streaks overlap spatially and temporally, producing accumulation effects that resemble thin fog and reduce scene contrast. To simulate this phenomenon, we introduce a rain accumulation model:
\begin{equation}
O = \alpha \left( B + \sum_{t=1}^{s} \tilde{S}_t R \right) + (1-\alpha) A,
\end{equation}
where $\tilde{S}_t$ denotes rain streak groups with similar orientation and shape, $s$ is the number of rain groups, $A$ is the global atmospheric light, and $\alpha$ represents the transmission factor controlling haze like attenuation.

By adjusting the number of rain groups $s$, overlap levels, and transmission $\alpha$, we generate a continuous degradation spectrum from light rain to heavy rainfall.

\subsubsection{Fog Synthesis}

Fog degradation mainly arises from atmospheric scattering. Under the assumption of homogeneous fog and stable illumination, the imaging process can be approximated by the atmospheric scattering model~\cite{zhang2017hazerd, lin2025seeing, lin2025nighthaze, jin2023enhancing}:

\begin{equation}
I(x) = J(x) t(x) + A (1 - t(x)),
\end{equation}

where $I(x)$ is the observed foggy image, $J(x)$ is the clear background image, $A$ denotes the global atmospheric light, and $t(x)$ represents the scene transmission map.

The transmission is related to the scene depth $d(x)$ and scattering coefficient $\beta$:

\begin{equation}
t(x) = \exp(-\beta d(x)).
\end{equation}

As the depth increases, the direct component $J(x)t(x)$ attenuates while the atmospheric light component $A(1-t(x))$ becomes dominant, resulting in reduced contrast and degraded visibility.

In practice, the haze free image is first converted to linear RGB space, the fog image is synthesized using the above equations, and the result is converted back to sRGB to preserve realistic brightness distribution.

Fog severity is controlled by the scattering coefficient $\beta$, which is determined from meteorological visibility $R_m$:

\begin{equation}
\beta = -\frac{\ln(\epsilon)}{R_m},
\end{equation}

where $\epsilon$ denotes the perceptual contrast threshold. By varying $R_m$, we generate fog conditions ranging from light to dense fog.

\subsubsection{Snow Synthesis}

Snow degradation mainly results from particle occlusion, light scattering, and contrast reduction under dense snowfall. Compared with rain streaks, snow particles appear as irregular high intensity granular structures with varying scales and non-uniform spatial distributions.

We model snowy degradation as the combination of background imaging, snow particle layers, and atmospheric scattering~\cite{michaelis2019benchmarking}:

\begin{equation}
I(x) = \gamma J(x) + \sum_{k=1}^{n} S_k(x;\rho,\sigma),
\end{equation}

where $I(x)$ denotes the synthesized snowy image, $J(x)$ is the clean background image, and $S_k(x;\rho,\sigma)$ represents the $k$-th snow particle layer generated with particle density $\rho$ and particle scale distribution $\sigma$. 
The parameter $\gamma$ denotes the transmission coefficient that models the visibility attenuation caused by snowfall. 
By adjusting $\rho$, $\sigma$, and $\gamma$, different snowfall severities ranging from light snow to heavy snow can be simulated.

Each snow layer consists of randomly generated particles with varying scale, brightness, and transparency, producing localized occlusions while preserving partial background visibility. Snowfall severity is controlled by the number of particle layers $n$, particle scale distribution, and transmission attenuation $\gamma(x)$, enabling gradual degradation from light snowfall to heavy snow conditions.

\subsection{Adverse Weather Severity Parameter Settings}

To simulate different levels of environmental degradation, we control the severity of fog, rain, and snow through a set of physically interpretable parameters. The specific configurations for light, moderate, and heavy conditions are summarized in Table~\ref{tab:weather_params}. For fog synthesis, $A$ denotes the global atmospheric light, while $\beta$ represents the scattering coefficient that determines the strength of atmospheric attenuation. As $\beta$ increases, light scattering becomes stronger, leading to reduced visibility and more severe fog degradation. For rain synthesis, $s$ denotes the number of rain streak groups, which directly controls the density of rain streaks, while $\alpha$ represents the transmission factor that models the accumulation effect of dense rainfall. Larger $s$ values produce denser rain streaks, and smaller $\alpha$ values correspond to stronger attenuation, resulting in heavier rain appearance. For snow synthesis, $\rho$ denotes the particle density controlling the number of snow particles, $\sigma$ represents the particle scale distribution that determines the size variation of snowflakes, and $\gamma$ denotes the transmission coefficient that models visibility attenuation under snowfall. Increasing $\rho$ and $\sigma$ leads to denser and larger snow particles, while decreasing $\gamma$ further reduces scene contrast, producing progressively heavier snowfall effects.

\begin{table}[h]
\centering
\caption{Parameter settings for different adverse weather severities.}
\label{tab:weather_params}
\scriptsize
\setlength{\tabcolsep}{8pt}
\renewcommand{\arraystretch}{1.3}
\begin{tabular}{lccc}
\toprule
Weather & Light & Moderate & Heavy \\
\midrule
Fog  & $A{=}0.8,\beta{=}0.32$ & $A{=}0.8,\beta{=}1.2$ & $A{=}0.8,\beta{=}2.0$ \\
Rain & $s{=}200,\alpha{=}0.9$ & $s{=}700,\alpha{=}0.85$ & $s{=}1000,\alpha{=}0.8$ \\
Snow & $\rho{=}0.50,\sigma{=}3.0,\gamma{=}0.80$ 
     & $\rho{=}0.85,\sigma{=}4.0,\gamma{=}0.70$ 
     & $\rho{=}0.90,\sigma{=}4.5,\gamma{=}0.55$ \\
\bottomrule
\end{tabular}
\end{table}

\section{Data Annotation of WeatherReasonSeg}

\begin{figure*}[t!]
    \centering
    \includegraphics[width=1\linewidth]{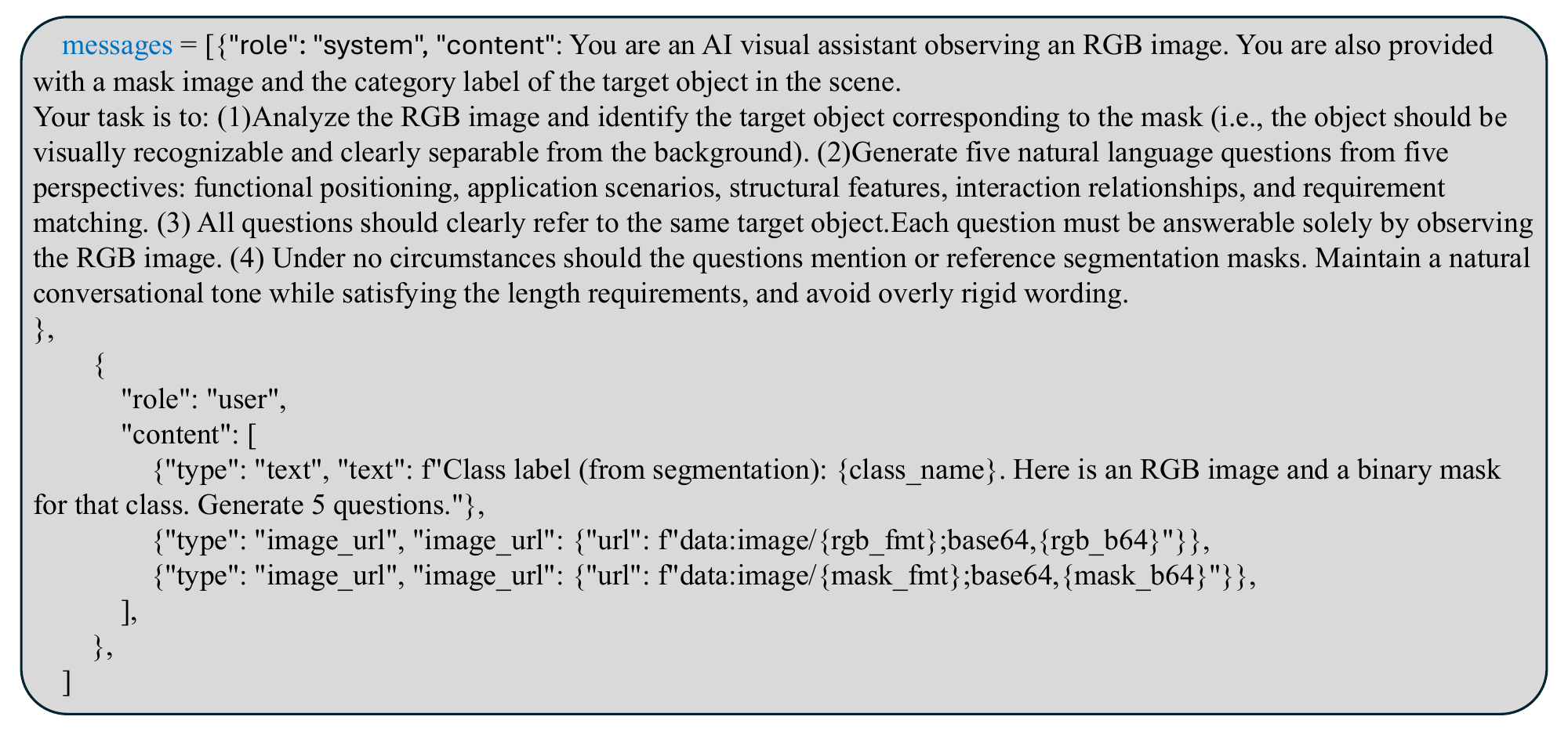}
    \caption{Prompt construction for reasoning query generation. The prompt consists of a system instruction defining the reasoning task and a user message containing the class label, RGB image, and corresponding binary mask to guide the generation of five reasoning oriented questions.}
    \label{fig_prompt}
\end{figure*}

\begin{figure*}[t!]
    \centering
    \includegraphics[width=1\linewidth]{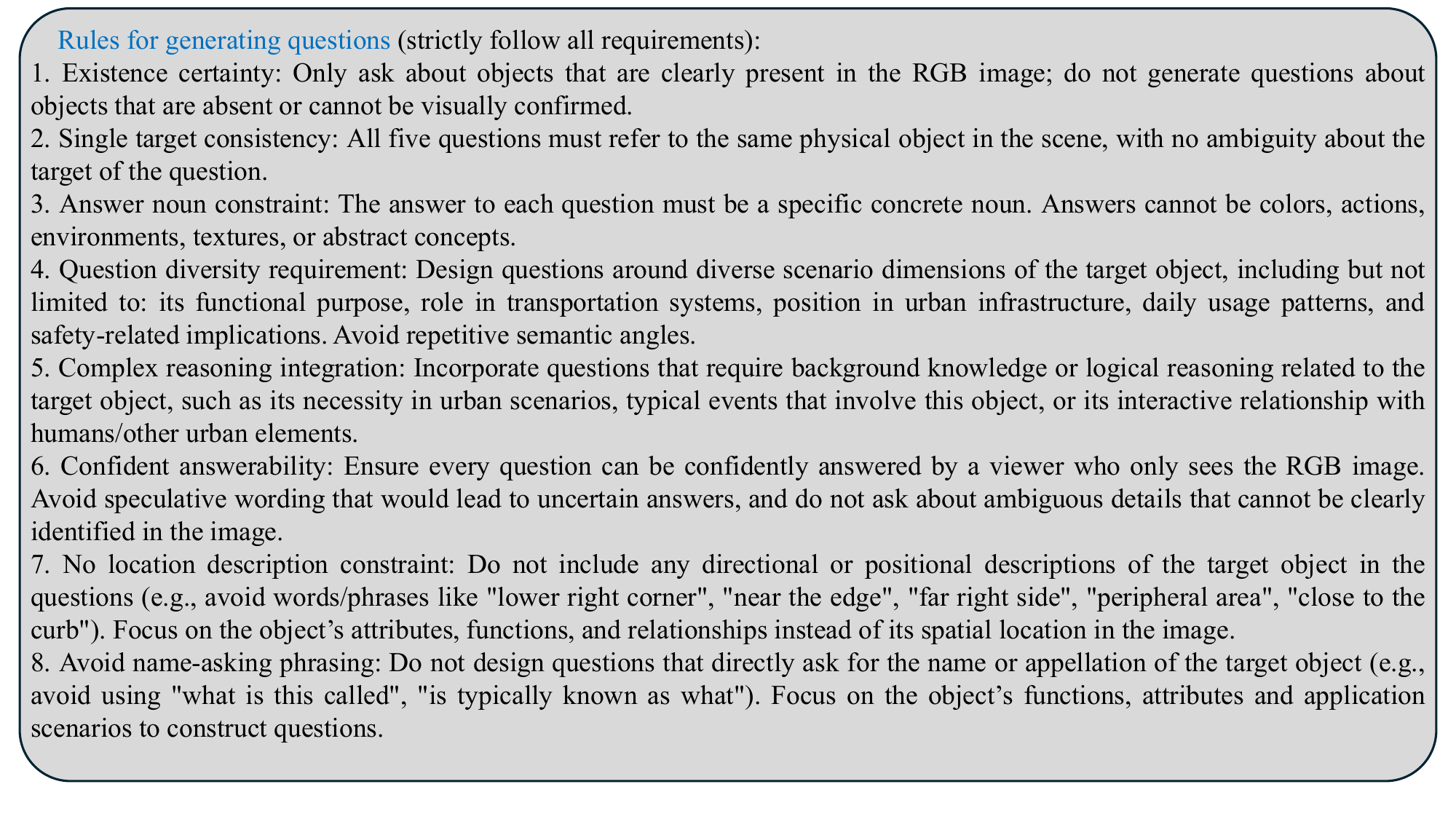}
    \caption{The figure outlines the verification rules used to ensure that generated questions satisfy object existence certainty, single target consistency, noun-based answers, reasoning diversity, and image grounded answer ability.}
    \label{fig_rules}
\end{figure*}

To construct the WeatherReasonSeg benchmark dataset, we propose an efficient semi-automatic data construction pipeline. Each sample in the dataset consists of an image, five reasoning queries, and a target binary mask. The overall process is described as follows.

\textbf{Step 1: Data Preparation and Pre-extraction} 
We first manually annotate 200 samples, covering queries and masks across all reasoning dimensions. Using a mask based extraction module, each image is processed at the pixel level to automatically decompose the scene into object masks and their corresponding descriptive labels.

\textbf{Step 2: Prompt Engineering with Large Language Models}
The extracted descriptive labels are incorporated as soft prompts, together with the original image masks and manually annotated example samples, to construct the complete prompt input for the Qwen large language model. Through an in context learning mechanism, the model generates five candidate reasoning queries for each sample.

\textbf{Step 3: Verification and Iteration} 
A dual stage filtering mechanism combining human experts and large language models is applied to verify the generated queries. Candidate queries are screened according to predefined rules. If semantic ambiguity or logical inconsistency is detected, the queries are either rewritten by the LLM or manually corrected by experts until high quality reasoning queries are obtained.

The prompt construction process in Step 2 is illustrated in Figure~\ref{fig_prompt}, while the rule based filtering procedure used for verification in Step 3 is illustrated in Figure~\ref{fig_rules}.

\section{Implementation details}
We adopt 8 NVIDIA 24G 3090 GPUs for inference. We evaluate four representative frameworks on WeatherReasonSeg: Grounded-SAM~\cite{ren2024grounded}, which is grounding-based, and three reasoning-based models: LISA~\cite{lai2024lisa}, Seg-R1~\cite{you2025seg}, and Seg-Zero~\cite{liu2025seg}. Specifically, LISA adopts a vision-language backbone based on LLaVA-1.5, while Seg-R1 and Seg-Zero are built upon the Qwen2.5-VL architecture. For fairness, all models are evaluated using their official implementations and default inference configurations. During evaluation, the input image and the corresponding query are provided to each framework to predict the segmentation mask of the target object. 

\section{More experiments}

\subsection{Qualitative Analysis Across Reasoning Dimensions}

\begin{figure*}[t!]
    \centering
    \includegraphics[width=1\linewidth]{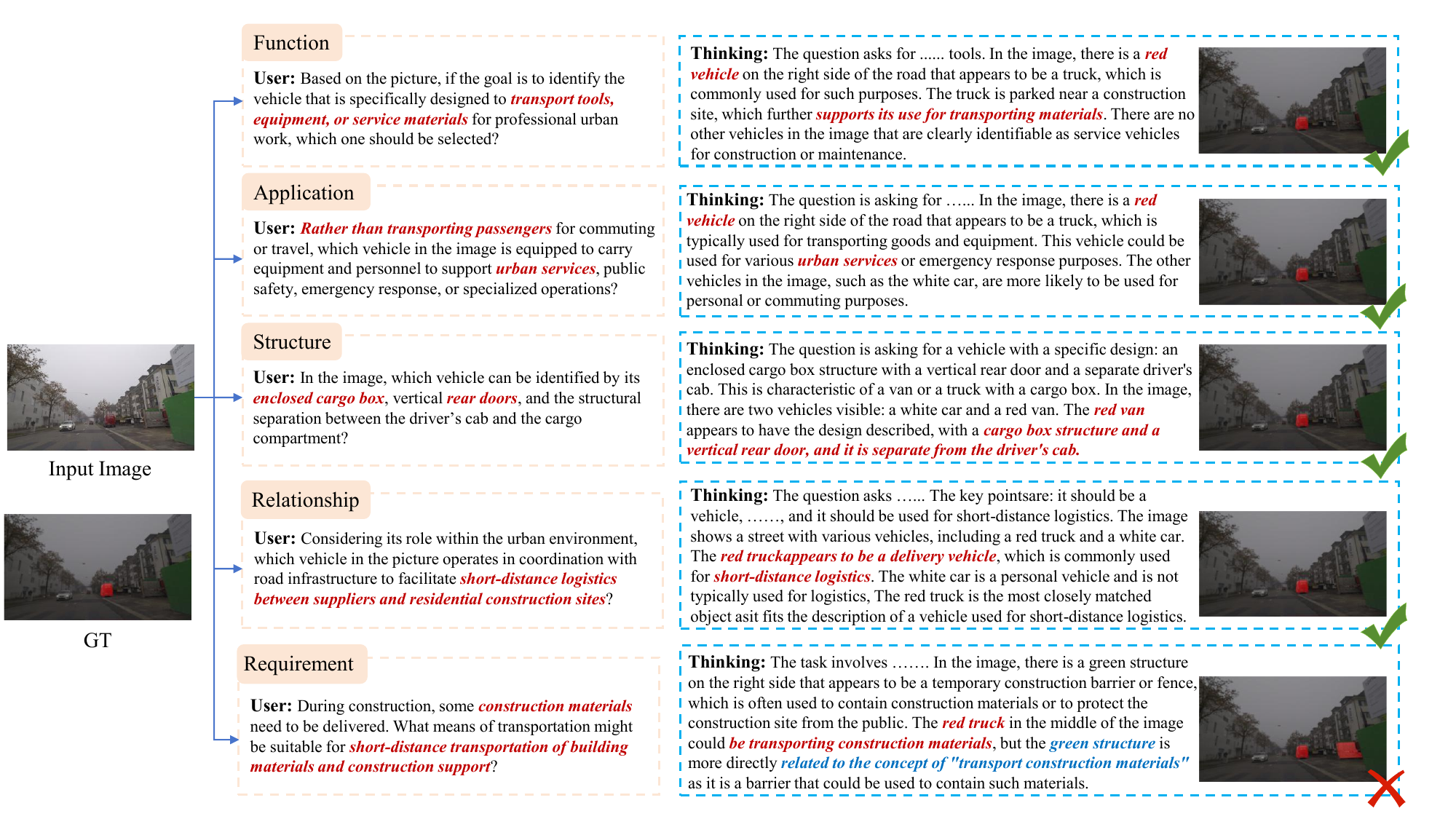}
    \caption{Given the same input image and target object, we construct queries from five semantic perspectives—Function, Application, Structure, Relationship, and Requirement. The reasoning segmentation method adopts Seg-Zero. The right panel illustrates the reasoning process and segmentation predictions of the VLM model, where green check marks denote correct predictions and the red cross indicates a failure case. This example highlights how different reasoning dimensions influence the model’s ability to correctly localize the target object.}
    \label{fig_dimension01}
\end{figure*}

\begin{figure*}[t!]
    \centering
    \includegraphics[width=1\linewidth]{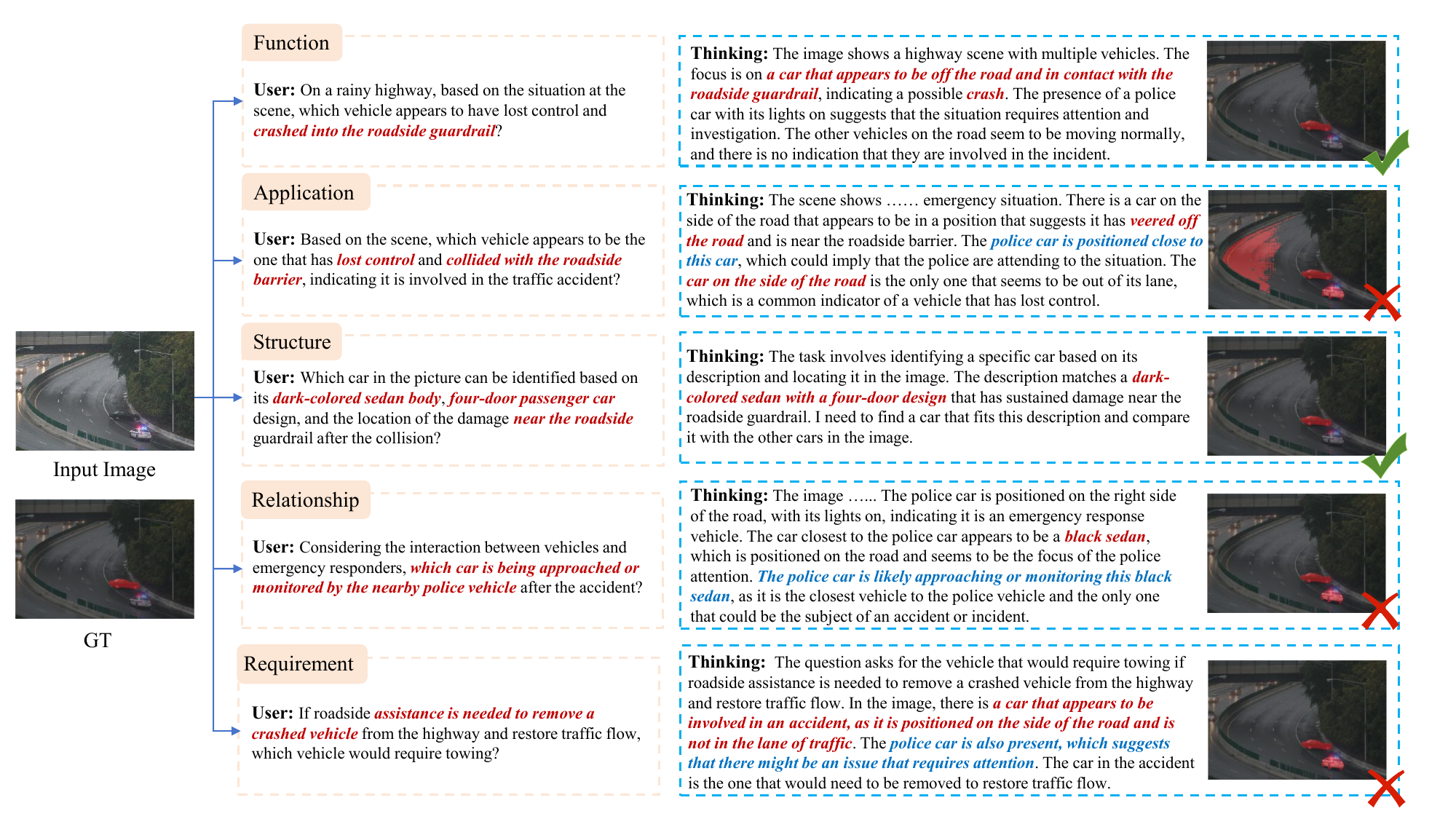}
    \caption{Given the same input image and target object, we test the reasoning and segmentation prediction results of the VLM model from five queries, where a green check indicates correct prediction and a red cross indicates prediction failure. This example highlights how different reasoning dimensions affect the model's ability to correctly locate the target object.}
    \label{fig_d2}
\end{figure*}

Figure~\ref{fig_dimension01} shows representative qualitative examples from our benchmark under real-world conditions, illustrating how reasoning queries from different semantic dimensions affect VLM reasoning segmentation. In this example, the target is a service vehicle, and queries are constructed from five perspectives: Function, Application, Structure, Relationship, and Requirement. The model correctly identifies the vehicle under Function, Application, Structure, and Relationship queries, indicating that when reasoning cues align with observable visual attributes, such as object category, structural characteristics, or roles in urban logistics, VLMs reasoning segmentation can reliably localize the target. However, the model fails under the Requirement query, confusing the functional roles of objects in the construction scene. Although the query asks for a vehicle suitable for transporting construction materials, the model incorrectly selects the temporary construction barrier. This error suggests that the model relies on contextual co-occurrence cues rather than correctly reasoning about object functions.

Figure~\ref{fig_d2} further reveals limitations under complex situational contexts. In the rainy highway accident scenario, the model successfully identifies the crashed vehicle when queries describe the accident event or structural characteristics, which provide clear visual grounding. In contrast, the model fails under certain Application and Relationship queries that require higher level situational reasoning, such as identifying the vehicle monitored by police or interpreting emergency response dynamics. In these cases, the model confuses the crashed vehicle with the police car.

Overall, the performance of reasoning segmentation varies significantly across different reasoning dimensions. Queries grounded in directly observable visual features tend to produce more stable predictions, whereas those requiring contextual reasoning, event interpretation, or implicit task requirements are more prone to failure. This observation is consistent with our quantitative findings and highlights the importance of evaluating reasoning segmentation across diverse semantic perspectives and adverse weather conditions.

\subsection{Qualitative Comparison Under Adverse Weather Conditions}

\begin{figure*}[t!]
    \centering
    \includegraphics[width=1\linewidth]{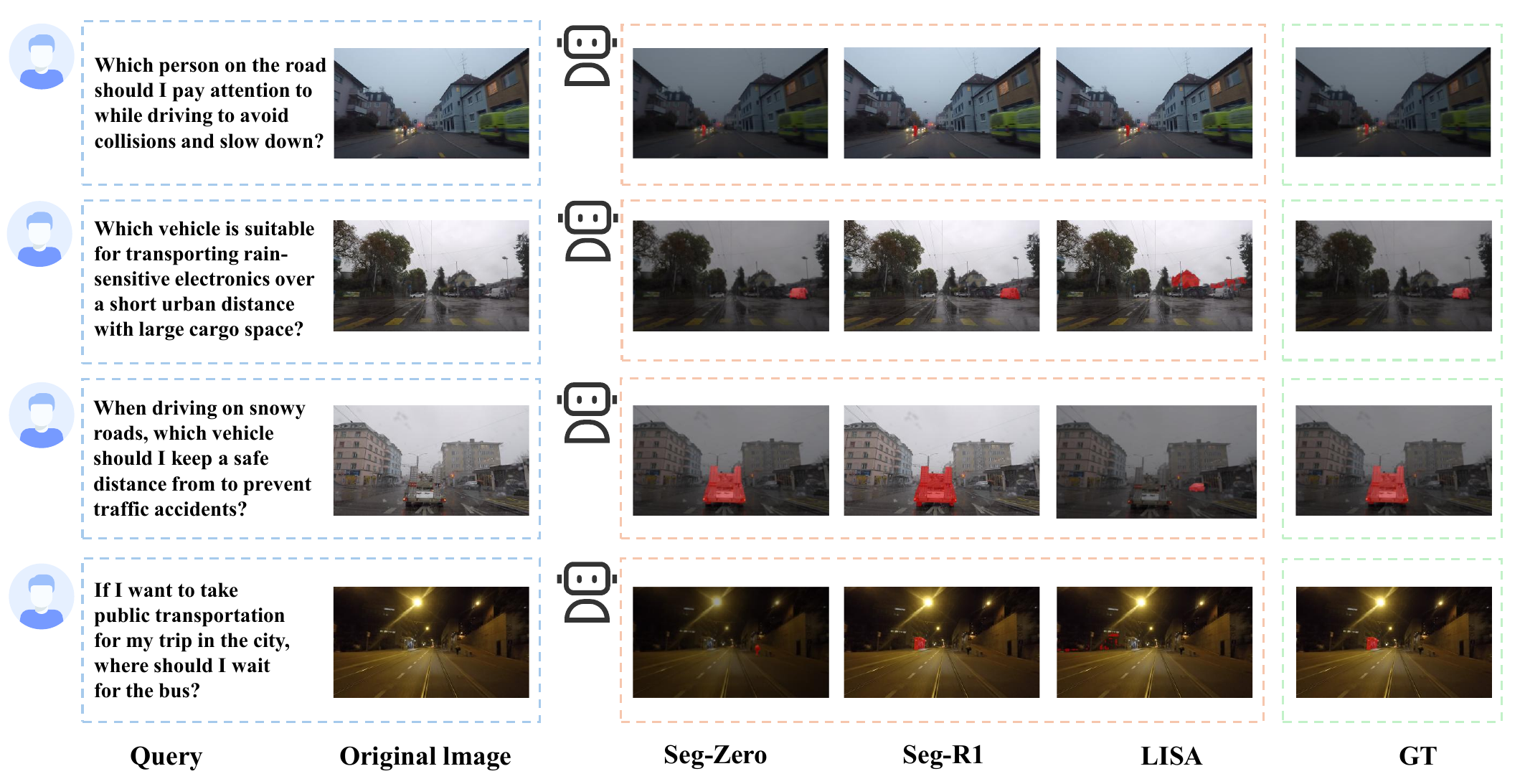}
    \caption{Qualitative results of reasoning segmentation under adverse weather conditions. For each example, we show the query, the original image, and the segmentation predictions produced by Seg-Zero, Seg-R1, and LISA, together with the ground truth (GT). The results highlight the different reasoning segmentation performance of VLM based methods in environments such as rain, snow, and nighttime.}
    \label{fig_real-ex}
\end{figure*}

Figure~\ref{fig_real-ex} shows the performance of reasoning based segmentation models degrades significantly under adverse weather conditions, consistent with the quantitative observations. When environmental degradation interferes with visual cues, reasoning based methods often struggle to accurately identify targets. For example, in rainy or snowy weather, Seg-Zero and Seg-R1 sometimes produce incomplete or off target predictions, while LISA may incorrectly highlight surrounding objects with similar contextual cues. These errors indicate that adverse weather not only affects low level visual perception but also interferes with the high level reasoning processes required for language guided segmentation. In contrast, truth masks demonstrate that accurate localization typically relies on the integration of object level features and contextual understanding, which remains a challenge for current VLM based reasoning based segmentation methods.

\section{Limitations}
Although WeatherReasonSeg provides a comprehensive benchmark for evaluating VLM reasoning segmentation under adverse weather conditions, it still has several limitations. First, the synthetic weather data are generated using simulation pipelines, which may not fully capture the complex physical properties of real-world weather phenomena. As a result, certain factors such as dynamic precipitation patterns, lighting interactions, and sensor noise are only partially modeled. Second, although our real-world subset includes multiple adverse conditions such as rain, snow, fog, and nighttime scenes, the diversity of environmental conditions and object categories remains limited compared with the vast variety of real-world scenarios ~\cite{li2024nightcc, li2026bridging, lin2025rgb, lin2025geocomplete, li2022mimt}. In addition, current evaluations focus primarily on instance-level reasoning and segmentation, and do not explicitly consider the broader hypothesis space of environmental and causal factors~\cite{chen2025hypospace}. 
Moreover, physically informative cues such as polarization, which have been shown to provide complementary robustness under degraded imaging conditions~\cite{yuan2025tailored, yuan2025mesa, yuan2025enhanced,yuan2025largescale, yuan2025engineering}, are not considered in the current benchmark.
Expanding the dataset to incorporate more physically accurate weather simulations and a richer diversity of real-world adverse environments is an important direction for future work.

\end{document}

%% file: preamble.tex
\usepackage[normalem]{ulem}